\documentclass{article}
\usepackage{spconf,amsmath,graphicx}
\usepackage{makecell}
\usepackage{amssymb}
\usepackage{graphics} 
\usepackage{epsfig} 
\usepackage{color}
\usepackage{makecell}
\usepackage{multirow}
\usepackage{color}
\usepackage{upgreek}
\usepackage{subfigure}
\usepackage{booktabs}
\usepackage[linesnumbered, ruled]{algorithm2e}

\DeclareMathOperator*{\argmin}{arg\,min}

\title{Cluster Regularized Quantization for Deep Networks Compression}
\name{Yiming Hu, Jianquan Li, Xianlei Long, Shenhua Hu, Jiagang Zhu, Xingang Wang, Qingyi Gu}
\address{Research Center of Precision Sensing and Control, Institute of Automation, Chinese Academy of Sciences\\ School of Computer and Control Engineering, University of Chinese Academy of Sciences\\
\{huyiming2016, qingyi.gu\}@ia.ac.cn}

\begin{document}
%
\maketitle
\begin{abstract}
 Deep neural networks (DNNs) have achieved great success in a wide range of computer vision areas, but the applications to mobile devices is limited due to their high storage and computational cost. Much efforts have been devoted to compress DNNs. In this paper, we propose a simple yet effective method for deep networks compression, named Cluster Regularized Quantization (CRQ),  which can reduce the presentation precision of a full-precision model to ternary values without significant accuracy drop. In particular, the proposed method aims at reducing the quantization error by introducing a cluster regularization term, which is imposed on the full-precision weights to enable them naturally concentrate around the target values. Through explicitly regularizing the weights during the re-training stage, the full-precision model can achieve the smooth transition to the low-bit one. Comprehensive experiments on benchmark datasets demonstrate the effectiveness of the proposed method.

\end{abstract}
\section{Introduction}
\label{sec:intro}

In the past few years, DNNs have been widely applied to various computer vision tasks, e.g., image classification, object detection, action recognition. However, it is very difficult to deploy these deep models on resource-constrained mobile devices due to their high storage and computational cost.

In order to make DNNs available on resource-constrained devices, many model compression and acceleration methods have been proposed including weight pruning \cite{Han2015, Hu2016}, low-rank approximation \cite{Denton2014, Wang2017}, and efficient model designs \cite{Howard2017,Zhang2017}. Another line of works \cite{WangP2018, Park2017} for network compression is to reduce the presentation precision of the full-precision parameters. Going one step further, restricting weights to some values with special types (e.g., zero or powers of two) can achieve fast inference on embedded devices such as the Field Programmable Gate Array (FPGA). Existing methods \cite{courbariaux2016binarized, Courbariaux2015} usually quantize weights by directly approximating the full-precision weights with the low-bit ones. XNOR-Net \cite{Rastegari2016} and TWN \cite{Feng2016Ternary} minimize the reconstruction error of the outputs between the full-precision model and the quantized one. These methods suffer from a common limitation: the reconstruction error of weights or intermediate outputs do not always truly reflect the change of the classification loss. Thus, directly cutting or sharing weights according to the reconstruction error of a layer may cause serious quantization error, which would bring significant disturbance to the classification loss and hurt the prediction accuracy of the quantized model.

To address the problem, we introduce an cluster regularization term, through which the full-precision weights can naturally concentrate around the target values during the re-training stage. After this, weights of each layer are quantized into corresponding cluster centers. Consequently, the quantization error can be significantly reduced. 

Our main contributions are summarized as follows. First, we present a new quantization method for deep networks compression by the introduction of the cluster regularization. Second, comprehensive experiments on benchmark datasets show that the proposed method can achieve competitive performance in accuracy and the rate of convergence.

\section{Related work}

The goal of network quantization is to replace the full-resolution weights with low-bit ones without significant accuracy drop.  This line of works can be traced back to two representative methods: Expectation Back-Propagation (EBP) \cite{Soudry2014} and BinaryConnect \cite{Courbariaux2015} in which authors explicitly constrained the weights to either $+1$ or $-1$ during propagations. These two methods achieve decent performance on small datasets including MNIST and CIFAR-10, yet they suffer from significant accuracy drop on challenging datasets such as ILSVRC-12 \cite{Russakovsky2015}. Afterwards, plenty of works were proposed to improve the performance of BinaryConnect. Rastegari et al. presented XNOR-Network \cite{Rastegari2016} that approximated the full-precision weights by introducing a scaling factor during binarization. For pursuing higher accuracy, High-Order Residual Quantization (HORQ)~{\cite{Li2017}} sought to compensate the information loss of binary quantization by conducting convolutional operations on inputs in different scales and then combined the results. The Ternary Weight Network (TWN) \cite{Feng2016Ternary} introduced zero as a third quantized value and was the first method that achieved decent results on the ILSVRC-12 dataset. Trained Ternary Quantization (TTQ) \cite{Zhu2016} achieved the state-of-the-art performance by taking the scaling factors as trainable parameters. Different from these works, our approach considers further reducing the quantization error to bridge the accuracy gap by introducing the cluster regularization with little computational complexity.

\section{Our Method}
\label{sec:pagestyle}
For better description of the proposed method, some notations are given first. Considering a layer of the DNN model, $X$ represents the input tensor, $W$ denotes the weight of the layer and $W_{\mathcal{Q}}$ represents the quantized weights of the layer.

\subsection{Motivation}
Most existing low-bit quantization methods \cite{courbariaux2016binarized,Courbariaux2015,Rastegari2016,Feng2016Ternary} can be classified into two categories. The first category such as BinaryNet \cite{courbariaux2016binarized} and BinaryConnect \cite{Courbariaux2015} directly approximates the full-precision weights with the low-bit ones:
\begin{equation}
\min_{W_{\mathcal{Q}}} \,\frac{1}{2N}||W- W_{\mathcal{Q}}||^2_2
\end{equation}
where $N$ is the size of $W$. The second category such as XNOR-Net \cite{Rastegari2016} and TWN \cite{Feng2016Ternary} usually minimizes the reconstruction error of a layer, which can be defined as the Euclidean distance of the outputs between the full-precision model and the quantized one:
\begin{equation}
\min_{W_{\mathcal{Q}}} \, \frac{1}{2N}||XW-XW_{\mathcal{Q}}||^2_2
\end{equation}

These methods suffer from a common limitation: the reconstruction error of weights or outputs between the full-precision model and the quantized one don't always truly reflect the change of the classification loss. Thus, directly cutting or sharing weights via the reconstruction error of a layer may cause serious quantization error, which would bring significant disturbance to the classification loss and hurt the prediction accuracy of the quantized model. This inspires us to define an additional regularization term, which is imposed on the weights of the full-precision model to enable them naturally concentrate around the target values by re-training. Particularly, the additional regularization can assist to reduce the quantization error and achieve the smooth transition from the full-precision model to the low-bit one.

\subsection{Formulation}
To softly modify the weights distribution by re-training, an cluster regularization term is defined. Han et al. \cite{HanS2015} apply $k$-means to assign weights with $K$ clusters layer-wise at first. Then the weights belonging to the same cluster centers share a same value. Unlike them, the values of cluster centers in our method are restricted to $T = \{-\alpha,0,+\alpha\}$, where $\alpha\in \mathbb{R}$ is a scaling factor. Without loss of generality, assuming $W$ is a row vector in $\mathbb{R}^{1\times N}$ and $M=[-1, 0, 1]^T$ is a column vector, where $N=c\times h \times z$, $c$, $h$ and $z$ represents the number of channels, the height and the width respectively. To partition weights $W$ into three clusters whose cluster centers are chosen from $T$, we solve the following optimization problem:

\begin{equation}
\begin{gathered}
J(Z, \alpha)	= \lVert W-\alpha M^TZ \lVert ^2 \\
\alpha^{*}, Z^{*} = \argmin \limits_{\alpha, Z} J(Z, \alpha) \\
s.t.\quad Z \in \{0, 1\}^{3\times N}, \sum_i^3 Z_{i,j} = 1, \forall j \in [1,N]
\end{gathered}
\label{eq:kmeans}
\end{equation}
where $Z \in \{0, 1\}^{3\times N}$ represents the cluster indicator, and the summation of each column is equal to $1$. It's worth noting that the optimization problem is different from the traditional $k$-means, since the cluster centers are restricted to the discrete values chosen from $T$. 

\subsection{Optimization Algorithm}
Eq. {\ref{eq:kmeans}} is non-convex, and optimization parameters involve $\alpha$ and $Z$. We consider using the alternating algorithm to solve this problem, i.e., solve $\alpha$ when $Z$ is fixed and vice versa. \\ \\
{\bfseries Solve $Z$ with $\alpha$ fixed.} By fixing $\alpha$, Eq. \ref{eq:kmeans} is written as:
\begin{equation}
\begin{gathered}
\min \limits_Z  {Tr(W^TW) - 2\alpha Tr(W^TM^TZ) + \alpha^2 Tr(Z^TMM^TZ)}\\
s.t.\quad Z \in \{0, 1\}^{3\times N}, \sum_i^3 Z_{i,j} = 1, \forall j \in [1,N]
\end{gathered}
\label{eq:solve_z}
\end{equation}
Directly solving $Z$ for Eq. \ref{eq:solve_z} is difficult due to its discrete constraints. Inspired by \cite{Shen2015}, $Z$ can be optimized column-wise by fixing other columns. In this way, $Z$ can be solved in roughly square time by learning only one column each time.  Solving Eq. \ref{eq:solve_z} is equivalent to optimizing two following general terms, i.e., $Tr(W^TM^TZ)$ and $Tr(Z^TMM^TZ)$. Let $w$ be the $i^{th}$ column of $W$ and $\tilde{W}$ indicates the vector of $W$ excluding $w$, $i = 1,2,\cdots,N$. Similarly, $z$ denotes the $i^{th}$ column of $Z$ and $\tilde{Z}$ represents the matrix of $Z$ excluding $z$. Then the following transformation can be got:

\begin{equation}
\begin{aligned}
Tr(W^TM^TZ) &=Tr((zw^T+\tilde{Z}\tilde{W}^{T})M^T)\\
&= C + z^TMw 
\end{aligned}
\label{eq:Tr_WTMZ}
\end{equation}
Here, those terms that don't contain $z$ are set to a constant $C$. Similarly, we have:

\begin{equation}
\begin{aligned}
Tr(Z^TMM^TZ)&=Tr(MM^T(zz^T+\tilde{Z}\tilde{Z}^T))\\
&= C + M^Tzz^TM 
\end{aligned}
\label{eq:Tr_ZTMMTZ}
\end{equation}
According to Eq. {\ref{eq:Tr_WTMZ}} and Eq. {\ref{eq:Tr_ZTMMTZ}}, Eq. {\ref{eq:solve_z}} can be written as follows:

\begin{equation}
\begin{gathered}
\min \limits_{q} \alpha^2q^2 - 2\alpha wq + C \\
s.t.\quad z \in \{0, 1\}^{3\times 1}, \sum_i^3 z_i = 1
\end{gathered}
\label{eq:SquareFunc}
\end{equation}
where $q = M^Tz$ and $q, \alpha, w \in \mathbb{R}$. Assuming Eq. \ref{eq:SquareFunc} is convex, it has the optimal solution when $q$ is equal to $\frac{w}{\alpha}$. But for this problem, $q$ is discrete. Let $H(M,\frac{w}{\alpha})$ returns the index of the nearest value of $M$ to $\frac{w}{\alpha}$, the optimal $z$ is solved as:

\begin{equation}
z_i=\left\{
\begin{aligned}
1 & \quad i = H(M,\frac{w}{\alpha}) \\
0 & \quad otherwise
\end{aligned}
\right.
\label{eq:solution_z}
\end{equation}
Each column of $Z$ is optimized according to Eq. {\ref{eq:solution_z}}. Then $Z$ can be solved after $N$ iterations.\\\\
{\bfseries Solve $\alpha$ with $Z$ fixed.} Eq. {\ref{eq:solve_z}} is convex when $Z$ is fixed. Eq. {\ref{eq:solve_z}} has the optimal solution when its derivative w.r.t. $\alpha$ is equal to zero. Thus we have:

\begin{equation}
\alpha = \frac{WZ^TM}{M^TZZ^TM}
\label{eq:solution_beta}
\end{equation}
In this way, $\alpha$ and $Z$ are updated iteratively until convergence. Here, it's not hard to find that solving $\alpha$ via Eq. \ref{eq:solution_beta} is equivalent to computing $\alpha$ as $\frac{1}{|I|}\sum_{i\in I} |W_i|$, where $I$ is the index sets of weights belonging to the cluster centers $-\alpha$ and $+\alpha$. Thus, solving Eq. \ref{eq:kmeans} using the alternating algorithm brings little computational complexity. 

As mentioned above, Eq. \ref{eq:kmeans} is used as a regularization term imposed on $W$ during the re-training stage. Assuming $L(W)$ is the loss function of the full-precision model, the objective function of the re-training stage is defined as follows:

\begin{equation}
\begin{gathered}
\min \limits_{W,Z,\alpha} L(W) + \lambda J(Z,\alpha)\\
s.t.\quad Z \in \{0, 1\}^{3\times N}, \sum_i^3 Z_{i,j} = 1, \forall j \in [1,N]
\end{gathered}
\label{loss_function}
\end{equation}
where $\lambda$ is a positive coefficient balancing the regularization term. Eq. \ref{loss_function} can be solved by iteratively updating $W$, $Z$ and $\alpha$. First, $Z$ and $\alpha$ are solved according to Eq. {\ref{eq:solve_z}} - {\ref{eq:solution_beta}}. Then $W$ is updated by stochastic gradient descent (SGD). We can derive the following weight update method:

\begin{algorithm} [!ht]
	\caption{Cluster Regularized Quantization}  
	\label{QuantizationPipeline}
	\SetKwInOut{Input}{Input}\SetKwInOut{Output}{Output}
	\Input {the training set $\mathcal{X}$ with size of $N$, the full precision weights $\{W_l:0<l<L\}$}
	\Output{ \{$\tilde{W_l}:0<l<L$\}: the quantized weights}
	
	\For{$n=1;n \le N;$}
	{
		\For{$l=1;l \le L;$}
		{
			Initialize $t=0$ and $\alpha = mean(|W_l|)$
			
			\Repeat{Convergence}{
				Compute $Z^t$ according to Equation \ref{eq:solve_z} - \ref{eq:solution_z}\\
				Compute $\alpha^t$ according to Equation \ref{eq:solution_beta}\\
				$t=t+1$
			}
			Update $W_l$ by the SGD step in Equation \ref{eq:weight_update}
		}
	}
	
	Quantize and fine-tune the re-trained model.
\end{algorithm}
\begin{equation}
W^t = W^{t-1} - \eta(\frac{\partial L}{\partial W^{t-1}} + \lambda(W^{t-1}-\alpha M^TZ))
\label{eq:weight_update}
\end{equation}
where $\eta$ is the learning rate. By the interaction among $Z$, $\alpha$ and $W$ during the re-training stage, the proposed regularization term enables $W$ naturally concentrate around the target values chosen from $T$.

\subsection{The Whole Quantization Framework}
After retraining, weights of each layer are quantized into corresponding cluster centers according to the indicator matrix $Z$ (Eq. \ref{eq:kmeans}), for parameter sharing. Specifically, the quantized weights are exactly equal to $\alpha M^T Z$ when Eq.\ref{eq:kmeans} has the optimal solution. In contrast, the re-training strategy with the cluster regularization brings less quantization error than the reconstruction-based methods \cite{Rastegari2016,Li2017}, since the weight $W$ remains in a highly clustered state after re-training. To further reduce the effects of quantization error to the classification loss, we fine-tune the re-trained model for several epochs. As in \cite{Courbariaux2015,Feng2016Ternary}, the quantized weights is used in the forward and backward process, while the full-precision weights are used in the stage of parameters update. The whole quantization framework is shown in Algorithm \ref{QuantizationPipeline}.

The pipeline of \cite{Wu2018, Ullrich2017} is close to ours, but we have to emphasize that our approach has three critical differences with them. First, we define and solve a new cluster optimization problem, in which the cluster centers are restricted to several discrete values. Solving the problem is very meaningful to the low-bit quantization, while the optimization algorithms in \cite{Wu2018, Ullrich2017} are not applicable to the special problem any more. Second, we seek to solve the cluster indicator $Z$ and the scaling factor $\alpha$ by the alternating algorithm, which avoids extremely time-consuming singular factorization operations. Third, our approach introduces fewer hyper-parameters and is easy to converge.

\section{Experiments}
\label{sec:typestyle}

In this section, we conduct comparison experiments on CIFAR-10 \cite{krizhevsky2009learning} and ImageNet \cite{Russakovsky2015}. The proposed CRQ is compared with two state-of-the-art methods including TWN \cite{Feng2016Ternary} and TTQ \cite{Zhu2016}. Our approach is implemented using the PyTorch \cite{paszkepytorch} framework. For the experiments on CIFAR-10 and ImageNet, we adopt the same settings as in \cite{Zhu2016}. Besides, $\lambda$ (Eq. \ref{loss_function}) is set to $0.001$.

\newcommand{\tabincell}[2]{\begin{tabular}{@{}#1@{}}#2\end{tabular}} 
\begin{table}
	\caption{Top-1 and Top-5 error rate (\%) of ternary AlexNet on ImageNet.}
	\begin{center}
		\begin{tabular}{c|cc|cc}
			\hline
			 Method & \tabincell{c}{Top-1} & \tabincell{c}{Top-1 $\downarrow$} & \tabincell{c}{Top-5} & \tabincell{c}{Top-5 $\downarrow$} \\
			\hline
			\hline
			Ref \cite{Zhu2016} & 42.80 & - 	& 	19.70 	& -	\\
			TWN \cite{Feng2016Ternary} & 45.50 & -2.70 & 23.20 & -3.50 \\
			TTQ \cite{Zhu2016} & 42.50 & 0.30 & 20.30 & -0.60 \\
			\hline
			Our	Ref	&	42.77	&	-	&	19.79	&	-\\
			Our CRQ & 42.02 & \textbf{0.75} & 19.18 & \textbf{0.61} \\
			\hline
		\end{tabular}
	\end{center}
	\label{table:TernaryAlexNet}
\end{table}

\subsection{ImageNet}
Table {\ref{table:TernaryAlexNet}} reports the comparison results of different methods on AlexNet. It can be observed that our ternary AlexNet can reach Top-1 error rate of $42.02\%$ and Top-5 error rate of $19.18\%$, which achieves the best performance compared with other methods and even outperforms its reference model by $0.75\%$ in Top-1 error rate. Next, further comparative experiments are conducted on ResNet models. Table {\ref{table:TernaryResNet}} summarizes the comparison results of different methods on ResNet-18. It's worth noting that our approach can still get a ternary ResNet-18 that has minor increase of the error rate: $1.99\%$ of Top-1 error rate and $1.38\%$ of Top-5 error rate, which are lower than TWN and TTQ.

\begin{figure}[t]
	\centering
	\includegraphics[width=0.80\columnwidth]{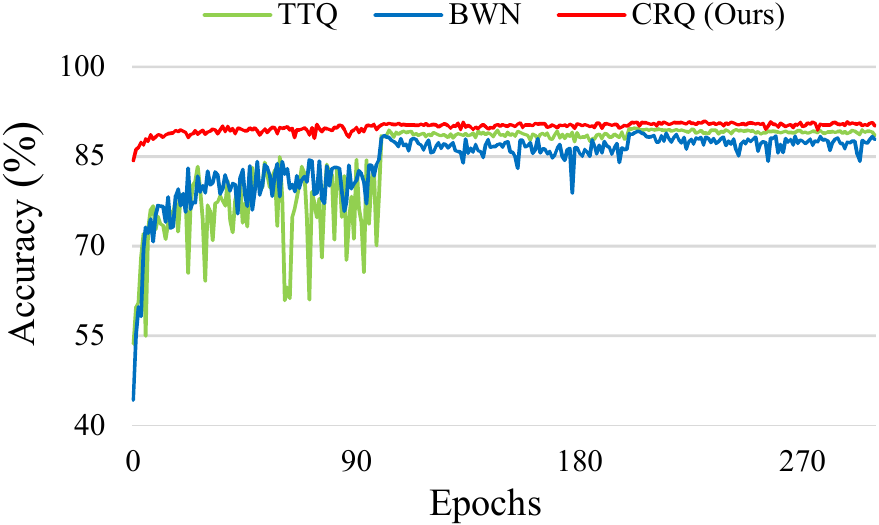}
	\vskip -2mm
	\caption{{Training accuracies (\%) of ternary ResNet-20 on CIFAR-10.}}
	\label{fig:Error_on_ImageNet}
\end{figure}

\subsection{CIFAR-10}

\begin{table}
	\caption{Top-1 and Top-5 error rate (\%) of ternary ResNet-18 on ImageNet.}
	\begin{center}
		\begin{tabular}{c|cc|cc}
			\hline
			Method & \tabincell{c}{Top-1} & \tabincell{c}{Top-1 $\downarrow$} & \tabincell{c}{Top-5} & \tabincell{c}{Top-5 $\downarrow$} \\
			\hline
			\hline
			Ref \cite{Zhu2016} &	30.40	&	-	&		10.80	&	-\\
			TWN \cite{Feng2016Ternary} & 34.70 & -4.30 & 13.80 & -3.00 \\
			TTQ \cite{Zhu2016}& 33.40 & -3.00 & 12.80 & -2.00 \\
			\hline
			Our Ref		&	30.89	&		-	&	11.18	&	-\\
			Our CRQ & 32.88 & \textbf{-1.99} & 12.56 & \textbf{-1.38} \\
			\hline
		\end{tabular}
	\end{center}
	\label{table:TernaryResNet}
\end{table}

\begin{table}
	\caption{Error rate (\%) of ternary ResNet-32 and ResNet-44 on CIFAR-10.}
	\begin{center}
		\begin{tabular}{c|cc|cc}
			\hline
			Model & Method & \tabincell{c}{Ref} & \tabincell{c}{Err.} & \tabincell{c}{Err. $\downarrow$} \\
			\hline
			\hline
			resnet-32 & TTQ \cite{Zhu2016} & 7.67 & 7.63 & 0.04 \\
			resnet-32 & CRQ & 7.74 & 7.61 & \textbf{0.13} \\
			\hline
			resnet-44 & TTQ & 7.18 & 7.02 & 0.16 \\
			resnet-44 & CRQ & 7.21 & 6.95 & \textbf{0.26} \\
			\hline
		\end{tabular}
	\end{center}
	\label{table:TernaryResNetCifar}
\end{table}

Table {\ref{table:TernaryResNetCifar}} shows the comparison results with TTQ on ResNet-32 and ResNet-44. Similar to the results on ImageNet, our method achieves bigger error rate drop than TTQ for the ternary ResNet-32 and ResNet-44. Particularly, our ternary ResNet-44 even outperforms the full-precision model by $0.26\%$ in the error rate, which may be resulted from the ability of the cluster regularization. Above experiments also imply that our CRQ can deal with the problem of ternary quantization with increased network depth.

\subsection{Convergence Analysis}
Going one step further, the rate of convergence of different methods is compared. We compare the evolution process of the training accuracy of different methods. Figure {\ref{fig:Error_on_ImageNet}} shows the training accuracy of ternary ResNet-20 changes over epochs on CIFAR-10. We can find that TWN, TTQ and our approach can converge to a minor error, but their accuracy curves are not as smooth as ours. Besides, our method converges within fewer epochs. These results demonstrate our approach has a faster rate of convergence.

\section{Conclusion}
\label{sec:print}

This paper proposes Cluster Regularized Quantization (CRQ), a ternary quantization method for deep networks compression. The key contribution is to define a new cluster regularization term, which significantly reduces the quantization error. Comprehensive experiments on CIFAR-10 and ImageNet demonstrate the CRQ can achieve competitive performance in accuracy and the rate of convergence.

\section{Acknowledgment}
This work has been supported by the National Key Research and Development Program of China No. 2018YFD0400902 and the National Natural Science
Foundation of China (Grant No. 61673376 and 61573349). 

{
\small
\bibliographystyle{IEEEbib}

}

\end{document}